\documentclass[technote,a4paper,leqno]{IEEEtran}
\pdfoutput=1

\usepackage[utf8]{inputenc} 
\usepackage[USenglish]{babel} 
\usepackage[T1]{fontenc}    
\usepackage{amsmath,amssymb}
\usepackage[absolute,overlay]{textpos}
\usepackage{tikz}
\usepackage{csquotes}
\usepackage[binary-units=true,detect-weight=true, detect-family=true]{siunitx}
\DeclareSIUnit\pixel{px}
\usepackage{caption}  

\usepackage{url}
\usepackage{breakurl}
\usepackage[raiselinks=true,
            bookmarks=true,
            bookmarksopenlevel=1,
            bookmarksopen=true,
            bookmarksnumbered=true,
            breaklinks,
            hyperindex=true,
            plainpages=false,
            pdfpagelabels=true,
            pdfborder={0 0 0.5}]{hyperref}

\usepackage{xspace}

\usepackage[nameinlink, noabbrev,capitalise]{cleveref}

\title{The HASYv2 dataset}
\author{%
    \IEEEauthorblockN{Martin Thoma}\\
    \IEEEauthorblockA{E-Mail: info@martin-thoma.de} 
}

\hypersetup{
    pdfauthor   = {Martin Thoma},
    pdfkeywords = {dataset},
    pdfsubject  = {HASY, HASYv2, dataset},
    pdftitle    = {The HASYv2 dataset},
}
\usepackage[inline]{enumitem}
\usepackage{longtable}
\usepackage{booktabs}       
\usepackage{braket}         
\usepackage{algorithm,algpseudocode}

\usepackage{mathtools}
\usepackage{mathrsfs}
\def\wasyfamily{\fontencoding{U}\fontfamily{wasy}\selectfont}
\usepackage{upgreek}
\usepackage{marvosym}
\usepackage{textcomp}
\usepackage{dsfont}
\usepackage{esint}
\usepackage{cmll}
\usepackage{stmaryrd}
\usepackage{skull}

\usepackage{parskip}
\usepackage{multirow}
\usepackage{microtype}

\newcommand{\dbTotalClasses}{369}
\newcommand{\dbTotalInstances}{\num{168233}}
\newcommand{\dbName}{HASY}
\newcommand{\dbNameVersion}{HASYv2}
\newcommand{\dbSizeMB}{34.6}
\newcommand{\dbDownloadURL}{\url{https://doi.org/10.5281/zenodo.259444}}
\newcommand{\dbMDfivesum}{fddf23f36e24b5236f6b3a0880c778e3}

\begin{document}
\maketitle
\begin{abstract}
This paper describes the \dbName{} dataset of \underline{ha}ndwritten
\underline{sy}mbols. \dbName{} is a publicly available,\footnote{See appendix for detailed instructions how to obtain the data.}
free of charge dataset of single symbols similar to MNIST. It contains
\dbTotalInstances~instances of \dbTotalClasses~classes. \dbName{} contains two
challenges: A classification challenge with 10~pre-defined folds for 10-fold
cross-validation and a verification challenge.
\end{abstract}

\section{Introduction}
Publicly available datasets have helped the computer vision community to
compare new algorithms and develop applications. Especially
MNIST~\cite{LeNet-5} was used thousands of times to train and evaluate models
for classification. However, even rather simple models consistently get about
$\SI{99.2}{\percent}$ accuracy on MNIST~\cite{TF-MNIST-2016}. The best models
classify everything except for about 20~instances correct. This makes
meaningful statements about improvements in classifiers hard. Possible reason
why current models are so good on MNIST are
\begin{enumerate*}
    \item MNIST has only 10~classes
    \item there are very few (probably none) labeling errors in MNIST
    \item every class has \num{6000}~training samples
    \item the feature dimensionality is comparatively low.
\end{enumerate*}
Also, applications which need to recognize only Arabic numerals are rare.

Similar to MNIST, \dbName{} is of very low resolution. In contrast to MNIST,
the \dbNameVersion~dataset contains \dbTotalClasses~classes, including Arabic
numerals and Latin characters. Furthermore, \dbNameVersion{} has much less
recordings per class than MNIST and is only in black and white whereas
MNIST is in grayscale.

\dbName{} could be used to train models for semantic segmentation of
non-cursive handwritten documents like mathematical notes or forms.

\section{Terminology}
A \textit{symbol} is an atomic semantic entity which has exactly one visual
appearance when it is handwritten. Examples of symbols are:
$\alpha, \propto, \cdot, x, \int, \sigma, \dots$

While a symbol is a single semantic entity with a given visual appearance, a
glyph is a single typesetting entity. Symbols, glyphs and \LaTeX{} commands do
not relate:

\begin{itemize}
    \item Two different symbols can have the same glyph. For example, the symbols
\verb+\sum+ and \verb+\Sigma+ both render to $\Sigma$, but they have different
semantics and hence they are different symbols.
    \item Two different glyphs can correspond to the same semantic entity. An example is
\verb+\varphi+ ($\varphi$) and \verb+\phi+ ($\phi$): Both represent the small
Greek letter \enquote{phi}, but they exist in two different variants. Hence
\verb+\varphi+ and \verb+\phi+ are two different symbols.
    \item Examples for different \LaTeX{} commands that represent the same symbol are
          \verb+\alpha+ ($\alpha$) and \verb+\upalpha+ ($\upalpha$): Both have the same
semantics and are hand-drawn the same way. This is the case for all \verb+\up+
variants of Greek letters.
\end{itemize}

All elements of the data set are called \textit{recordings} in the following.

\section{How HASY was created}
\dbName{} is derived from the HWRT dataset which was first used and described
in~\cite{Thoma:2014}. HWRT is an on-line recognition dataset, meaning it does
not contain the handwritten symbols as images, but as point-sequences. Hence
HWRT contains strictly more information than \dbName. The smaller dimension
of each recordings bounding box was scaled to be \SI{32}{\pixel}. The image
was then centered within the $\SI{32}{\pixel} \times \SI{32}{\pixel}$ bounding
box.

\begin{figure}[h]
    \centering
    \includegraphics*[width=\linewidth, keepaspectratio]{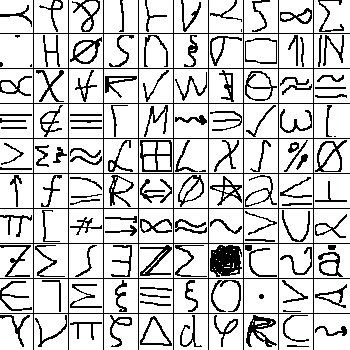}
    \caption{100 recordings of the \dbNameVersion{} data set.}
    \label{fig:100-data-items}
\end{figure}

HWRT contains exactly the same recordings and classes as \dbName, but \dbName{}
is rendered in order to make it easy to use.

HWRT and hence \dbName{} is a merged dataset. $\SI{91.93}{\percent}$ of HWRT
were collected by Detexify~\cite{Kirsch,Kirsch2014}. The remaining recordings
were collected by \href{http://write-math.com}{http://write-math.com}. Both
projects aim at helping users to find \LaTeX{} commands in cases where the
users know how to write the symbol, but not the symbols name. The user writes
the symbol on a blank canvas in the browser (either via touch devices or with a
mouse). Then the websites give the Top-$k$ results which the user could have
thought of. The user then clicks on the correct symbol to accept it as the
correct symbol. On \href{http://write-math.com}{write-math.com}, other users
can also suggest which symbol could be the correct one.

After collecting the data, Martin Thoma manually inspected each recording. This
manual inspection is a necessary step as anonymous web users could submit any
drawing they wanted to any symbol. This includes many creative recordings as
shown in~\cite{Kirsch,Thoma:2014} as well as loose associations. In some cases,
the correct label was unambiguous and could be changed. In other cases, the
recordings were removed from the data set.

It is not possible to determine the exact number of people who contributed
handwritten symbols to the Detexify part of the dataset. The part which was
created with \href{http://write-math.com}{write-math.com} was created by
477~user~IDs. Although user IDs are given in the dataset, they are not
reliable. On the one hand, the Detexify data has the user ID 16925,
although many users contributed to it. Also, some users lend their phone to
others while being logged in to show how write-math.com works. This leads to
multiple users per user ID. On the other hand, some users don't register and
use write-math.com multiple times. This can lead to multiple user IDs for one
person.

\section{Classes}
The \dbNameVersion~dataset contains \dbTotalClasses~classes. Those classes include the
Latin uppercase and lowercase characters (\verb+A-Z+, \verb+a-z+), the Arabic
numerals (\verb+0-9+), 32~different types of arrows, fractal and calligraphic
Latin characters, brackets and more. See \cref{table:symbols-of-db-0,table:symbols-of-db-1,table:symbols-of-db-2,table:symbols-of-db-3,table:symbols-of-db-4,table:symbols-of-db-5,table:symbols-of-db-6,table:symbols-of-db-7,table:symbols-of-db-8} for more information.

\section{Data}
The \dbNameVersion~dataset contains \dbTotalInstances{} black and white images
of the size $\SI{32}{\pixel} \times \SI{32}{\pixel}$. Each image is labeled
with one of \dbTotalClasses~labels. An example of 100~elements of the
\dbNameVersion{} data set is shown in~\cref{fig:100-data-items}.

The average amount of black pixels is \SI{16}{\percent}, but this is highly
class-dependent ranging from \SI{3.7}{\percent} of \enquote{$\dotsc$} to \SI{59.2}{\percent} of \enquote{$\blacksquare$} average
black pixel by class.

\begin{figure}[h]
    \centering
    \includegraphics*[width=\linewidth, keepaspectratio]{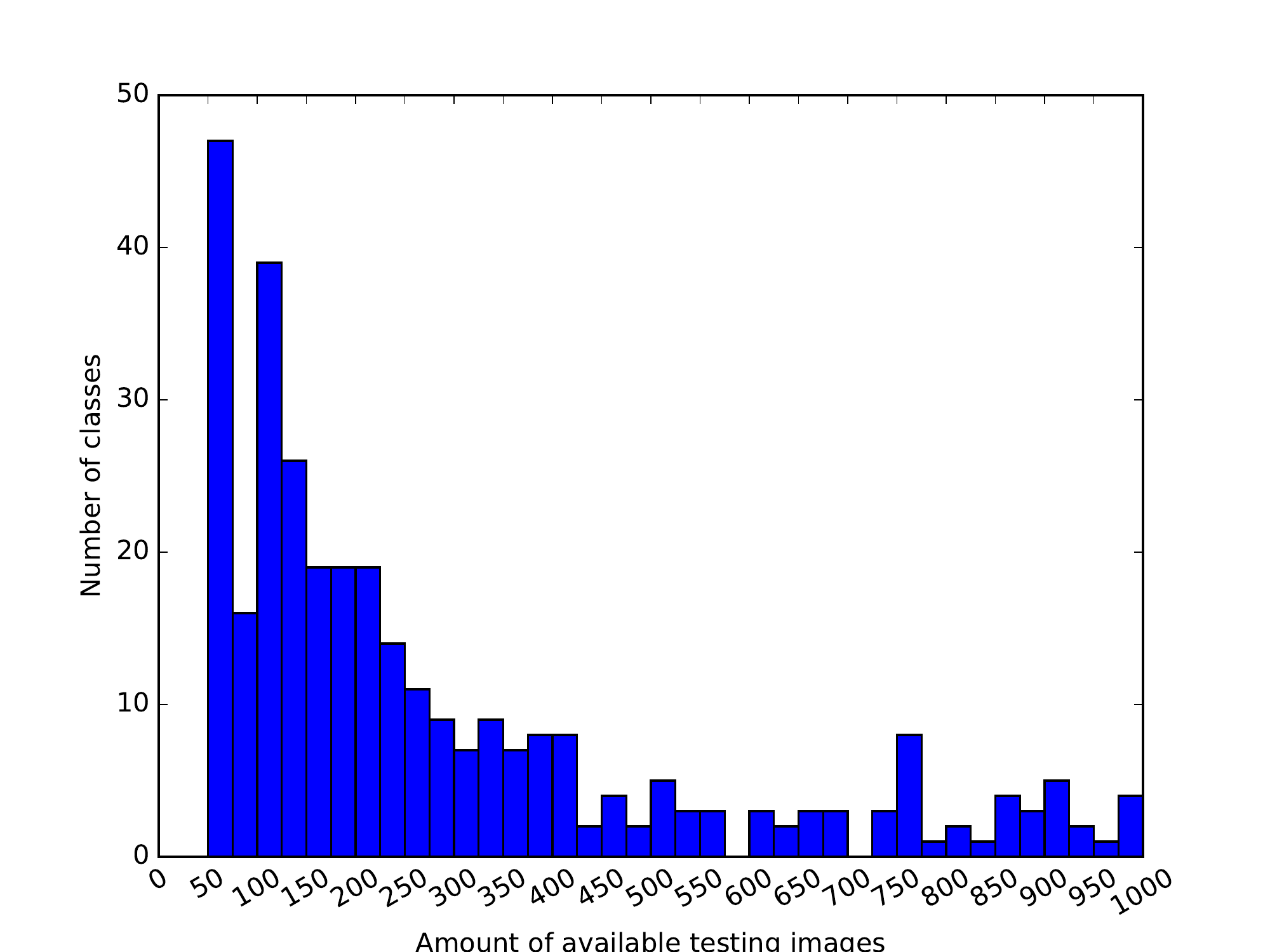}
    \caption{Distribution of the data among classes. 47~classes with
             more than \num{1000} recordings are not shown.}
    \label{fig:class-data-distribution}
\end{figure}

The ten classes with most recordings are:
\[\int, \sum, \infty, \alpha, \xi, \equiv, \partial, \mathds{R}, \in, \square\]
Those symbols have \num{26780} recordings and thus account for
\SI{16}{\percent} of the data set. 47~classes have more than \num{1000}
recordings. The number of recordings of the remaining classes are distributed
as visualized in~\cref{fig:class-data-distribution}.

A weakness of \dbNameVersion{} is the amount of available data per class. For
some classes, there are only 51~elements in the test set.

The data has $32\cdot 32 = 1024$ features in $\Set{0, 255}$.
As~\cref{table:pca-explained-variance} shows, \SI{32}{\percent} of the features
can explain~\SI{90}{\percent} of the variance, \SI{54}{\percent} of the
features explain \SI{99}{\percent} of the variance and \SI{86}{\percent} of the
features explain \SI{99}{\percent} of the variance.

\begin{table}[h]
    \centering
    \begin{tabular}{lccc}
    \toprule
    Principal Components &  331              & 551               & 882  \\
    Explained Variance   & \SI{90}{\percent} & \SI{95}{\percent} & \SI{99}{\percent} \\
    \bottomrule
    \end{tabular}
    \caption{The number of principal components necessary to explain,
             \SI{90}{\percent}, \SI{95}{\percent}, \SI{99}{\percent}
             of the data.}
    \label{table:pca-explained-variance}
\end{table}

\begin{figure}[h]
    \centering
    \includegraphics*[width=\linewidth, keepaspectratio]{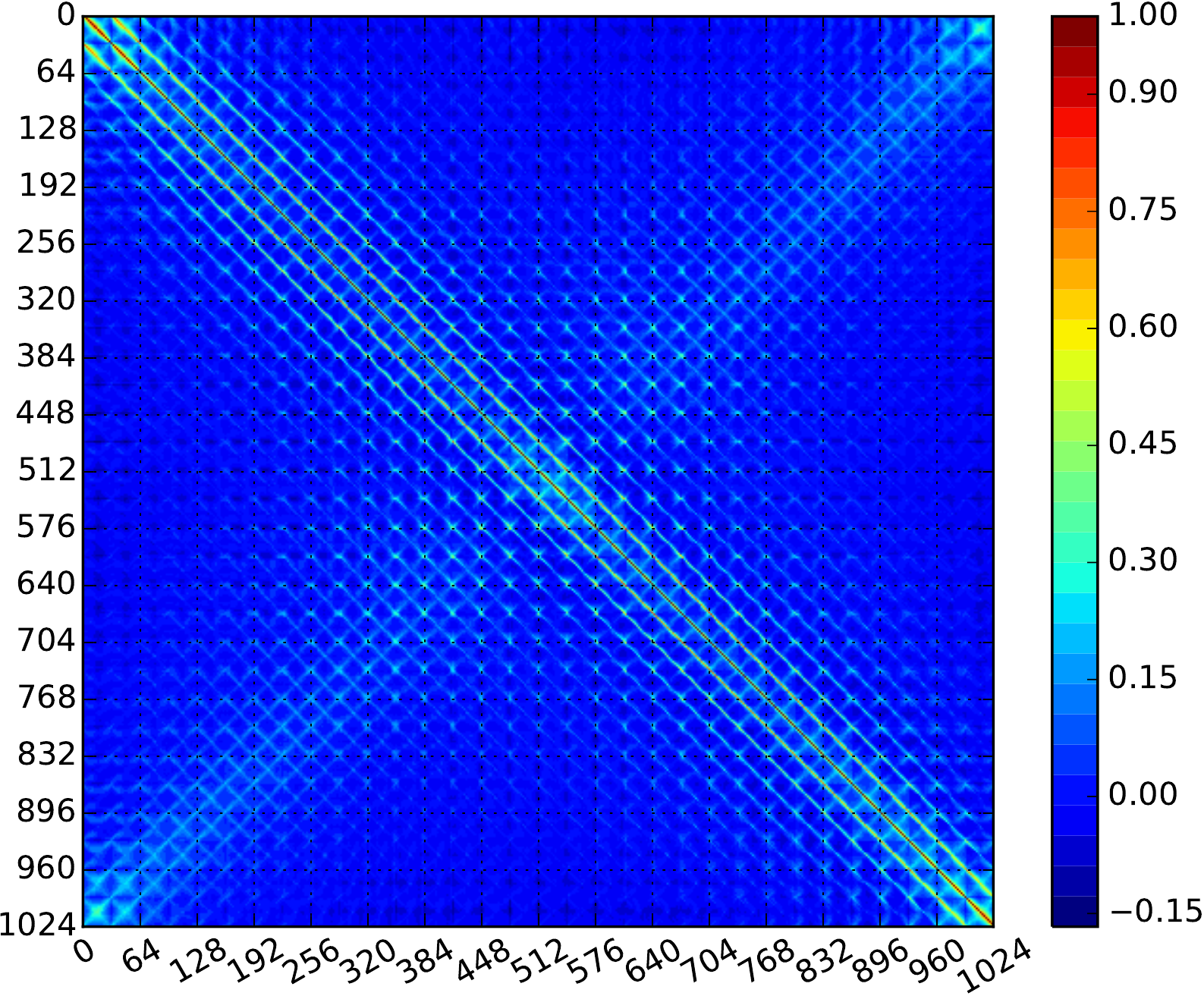}
    \caption{Correlation of all $32 \cdot 32 = 1024$ features. The diagonal
             shows the correlation of a feature with itself.}
    \label{fig:feature-correlation}
\end{figure}

The Pearson correlation coefficient was calculated for all features. The
features are more correlated the closer the pixels are together as one can see
in~\cref{fig:feature-correlation}. The block-like structure of every 32th
feature comes from the fact the features were flattened for this visualization.
The second diagonal to the right shows features which are one pixel down in the
image. Those correlations are expected as symbols are written by continuous
lines. Less easy to explain are the correlations between high-index
features with low-index features in the upper right corner of the image.
Those correlations correspond to features in the upper left corner with
features in the lower right corner. One explanation is that symbols which have
a line in the upper left corner are likely $\blacksquare$.

\section{Classification Challenge}
\subsection{Evaluation}
\dbName{} defines 10 folds which should be used for calculating the accuracy
of any classifier being evaluated on \dbName{} as follows:

\begin{algorithm}[H]
    \begin{algorithmic}
        \Function{CrossValidation}{Folds $F$}
            \State $D \gets \cup_{i=1}^{10} F_i$\Comment{Complete Dataset}
            \For{($i=0$; $\;i < 10$; $\;i$++)}
                \State $A \gets D \setminus F_i$\Comment{Train set}
                \State $B \gets F_i$\Comment{Test set}
                \State Train Classifier $C_i$ on $A$
                \State Calculate accuracy $a_i$ of $C_i$ on $B$
            \EndFor
            \State \Return ($\frac{1}{10}\sum_{i=1}^{10} a_i$, $\min(a_i)$, $\max(a_i)$)
        \EndFunction
    \end{algorithmic}
    \caption{Calculate the mean accuracy, the minimum accuracy, and the maximum
             accuracy with 10-fold cross-validation}
\label{alg:seq1}
\end{algorithm}

\subsection{Model Baselines}
Eight standard algorithms were evaluated by their accuracy on the raw image
data. The neural networks were implemented with
Tensorflow~\cite{tensorflow2015-whitepaper}. All other algorithms are
implemented in sklearn~\cite{scikit-learn}. \Cref{table:classifier-results}
shows the results of the models being trained and tested on MNIST and also for
\dbNameVersion{}:
\begin{table}[h]
    \centering
    \begin{tabular}{lrrr}
    \toprule
    \multirow{2}{*}{Classifier}    & \multicolumn{3}{c}{Test Accuracy}           \\
                  & MNIST                & HASY                & min -- max\hphantom{00 } \\\midrule
    TF-CNN        & \SI{99.20}{\percent} & \SI{81.0}{\percent} & \SI{80.6}{\percent} -- \SI{81.5}{\percent}\\
    Random Forest & \SI{96.41}{\percent} & \SI{62.4}{\percent} & \SI{62.1}{\percent} -- \SI{62.8}{\percent}\\
    MLP (1 Layer) & \SI{89.09}{\percent} & \SI{62.2}{\percent} & \SI{61.7}{\percent} -- \SI{62.9}{\percent}\\
    LDA           & \SI{86.42}{\percent} & \SI{46.8}{\percent} & \SI{46.3}{\percent} -- \SI{47.7}{\percent}\\
    QDA           & \SI{55.61}{\percent} & \SI{25.4}{\percent} & \SI{24.9}{\percent} -- \SI{26.2}{\percent}\\
    Decision Tree & \SI{65.40}{\percent} & \SI{11.0}{\percent} & \SI{10.4}{\percent} -- \SI{11.6}{\percent}\\
    Naive Bayes   & \SI{56.15}{\percent} &  \SI{8.3}{\percent} & \SI{7.9}{\percent} -- \hphantom{0}\SI{8.7}{\percent}\\
    AdaBoost      & \SI{73.67}{\percent} &  \SI{3.3}{\percent} & \SI{2.1}{\percent} -- \hphantom{0}\SI{3.9}{\percent}\\
    \bottomrule
    \end{tabular}
    \caption{Classification results for eight classifiers.
             The number of
             test samples differs between the folds, but is $\num{16827} \pm
             166$. The decision tree
             was trained with a maximum depth of 5. The exact structure
             of the CNNs is explained in~\cref{subsec:CNNs-Classification}.}
    \label{table:classifier-results}
\end{table}

The following observations are noteworthy:
\begin{itemize}
    \item All algorithms achieve much higher accuracy on MNIST than on
          \dbNameVersion{}.
    \item While a single Decision Tree performs much better on MNIST than
          QDA, it is exactly the other way around for~\dbName{}. One possible
          explanation is that MNIST has grayscale images while \dbName{} has
          black and white images.
\end{itemize}

\subsection{Convolutional Neural Networks}\label{subsec:CNNs-Classification}
Convolutional Neural Networks (CNNs) are state of the art on several computer
vision benchmarks like MNIST~\cite{wan2013regularization}, CIFAR-10, CIFAR-100
and SVHN~\cite{huang2016densely},
ImageNet~2012~\cite{deep-residual-networks-2015} and more. Experiments on
\dbNameVersion{} without preprocessing also showed that even the
simplest CNNs achieve much higher accuracy on \dbNameVersion{} than all other
classifiers (see~\cref{table:classifier-results}).

\Cref{table:cnn-results} shows the 10-fold cross-validation results for four
architectures.
\begin{table}[H]
    \centering
    \begin{tabular}{lrrrr}
    \toprule
    \multirow{2}{*}{Network} & \multirow{2}{*}{Parameters}    & \multicolumn{2}{c}{Test Accuracy} & \multirow{2}{*}{Time} \\
            &               & mean                & min -- max\hphantom{00 } & \\\midrule
    2-layer & \num{3023537} & \SI{73.8}{\percent} & \SI{72.9}{\percent} -- \SI{74.3}{\percent} & \SI{1.5}{\second}\\
    3-layer & \num{1530609} & \SI{78.4}{\percent} & \SI{77.6}{\percent} -- \SI{79.0}{\percent} & \SI{2.4}{\second}\\
    4-layer &  \num{848753} & \SI{80.5}{\percent} & \SI{79.2}{\percent} -- \SI{80.7}{\percent} & \SI{2.8}{\second}\\
    TF-CNN  & \num{4592369} & \SI{81.0}{\percent} & \SI{80.6}{\percent} -- \SI{81.5}{\percent} & \SI{2.9}{\second}\\
    \bottomrule
    \end{tabular}
    \caption{Classification results for CNN architectures. The test time is,
             as before, the mean test time for all examples on the ten folds.}
    \label{table:cnn-results}
\end{table}
The following architectures were evaluated:
\begin{itemize}
    \item 2-layer: A convolutional layer with 32~filters of size $3 \times 3 \times 1$
          is followed by a $2 \times 2$ max pooling layer with stride~2. The output
          layer is --- as in all explored CNN architectures --- a fully
          connected softmax layer with 369~neurons.
    \item 3-layer: Like the 2-layer CNN, but before the output layer is another
          convolutional layer with 64~filters of size $3 \times 3 \times 32$
          followed by a $2 \times 2$ max pooling layer with stride~2.
    \item 4-layer: Like the 3-layer CNN, but before the output layer is another
          convolutional layer with 128~filters of size $3 \times 3 \times 64$
          followed by a $2 \times 2$ max pooling layer with stride~2.
    \item TF-CNN: A convolutional layer with 32~filters of size $3 \times 3 \times 1$
          is followed by a $2 \times 2$ max pooling layer with stride~2.
          Another convolutional layer with 64~filters of size $3 \times 3 \times 32$
          and a $2 \times 2$  max pooling layer with stride~2 follow. A fully
          connected layer with 1024~units and tanh activation function, a
          dropout layer with dropout probability 0.5 and the output softmax
          layer are last. This network is described in~\cite{tf-mnist}.
\end{itemize}

For all architectures, ADAM~\cite{kingma2014adam} was used for training. The
combined training and testing time was always less than 6~hours for the 10~fold
cross-validation on a Nvidia GeForce GTX Titan Black with CUDA~8 and CuDNN~5.1.
\clearpage
\subsection{Class Difficulties}
The class-wise accuracy
\[\text{class-accuracy}(c) = \frac{\text{correctly predicted samples of class } c}{\text{total number of training samples of class } c}\]
is used to estimate how difficult a class is.

32~classes were not a single time classified correctly by TF-CNN and hence have
a class-accuracy of~0. They are shown in~\cref{table:hard-classes}. Some, like
\verb+\mathsection+ and \verb+\S+ are not distinguishable at all. Others, like
\verb+\Longrightarrow+ and
\verb+\Rightarrow+ are only distinguishable in some peoples handwriting.
Those classes account for \SI{2.8}{\percent} of the data.

\begin{table}[h]
    \centering
    \begin{tabular}{lcrlc}
    \toprule
    \LaTeX & Rendered & Total & Confused with & \\\midrule
    \verb+\mid+ & $\mid$ & 34 & \verb+|+ & $|$ \\
    \verb+\triangle+ & $\triangle$ & 32 & \verb+\Delta+ & $\Delta$ \\
    \verb+\mathds{1}+ & $\mathds{1}$ & 32 & \verb+\mathbb{1}+ & $\mathbb{1}$ \\
    \verb+\checked+ & {\mbox {\wasyfamily \char 8}} & 28 & \verb+\checkmark+ & $\checkmark$ \\
    \verb+\shortrightarrow+ & $\shortrightarrow$ & 28 & \verb+\rightarrow+ & $\rightarrow$ \\
    \verb+\Longrightarrow+ & $\Longrightarrow$ & 27 & \verb+\Rightarrow+ & $\Rightarrow$ \\
    \verb+\backslash+ & $\backslash$ & 26 & \verb+\setminus+ & $\setminus$ \\
    \verb+\O+ & \O & 24 & \verb+\emptyset+ & $\emptyset$ \\
    \verb+\with+ & $\with$ & 21 & \verb+\&+ & $\&$ \\
    \verb+\diameter+ & {\mbox {\wasyfamily \char 31}} & 20 & \verb+\emptyset+ & $\emptyset$ \\
    \verb+\triangledown+ & $\triangledown$ & 20 & \verb+\nabla+ & $\nabla$ \\
    \verb+\longmapsto+ & $\longmapsto$ & 19 & \verb+\mapsto+ & $\mapsto$ \\
    \verb+\dotsc+ & $\dotsc$ & 15 & \verb+\dots+ & $\dots$ \\
    \verb+\fullmoon+ & {\mbox {\wasyfamily \char 35}} & 15 & \verb+\circ+ & $\circ$ \\
    \verb+\varpropto+ & $\varpropto$ & 14 & \verb+\propto+ & $\propto$ \\
    \verb+\mathsection+ & $\mathsection$ & 13 & \verb+\S+ & $\S$ \\
    \verb+\vartriangle+ & $\vartriangle$ & 12 & \verb+\Delta+ & $\Delta$ \\
    \verb+O+ & $O$ & 9 & \verb+\circ+ & $\circ$ \\
    \verb+o+ & $o$ & 7 & \verb+\circ+ & $\circ$ \\
    \verb+c+ & $c$ & 7 & \verb+\subset+ & $\subset$ \\
    \verb+v+ & $v$ & 7 & \verb+\vee+ & $\vee$ \\
    \verb+x+ & $x$ & 7 & \verb+\times+ & $\times$ \\
    \verb+\mathbb{Z}+ & $\mathbb{Z}$ & 7 & \verb+\mathds{Z}+ & $\mathds{Z}$ \\
    \verb+T+ & $T$ & 6 & \verb+\top+ & $\top$ \\
    \verb+V+ & $V$ & 6 & \verb+\vee+ & $\vee$ \\
    \verb+g+ & $g$ & 6 & \verb+9+ & $9$ \\
    \verb+l+ & $l$ & 6 & \verb+|+ & $|$ \\
    \verb+s+ & $s$ & 6 & \verb+\mathcal{S}+ & $\mathcal{S}$ \\
    \verb+z+ & $z$ & 6 & \verb+\mathcal{Z}+ & $\mathcal{Z}$ \\
    \verb+\mathbb{R}+ & $\mathbb{R}$ & 6 & \verb+\mathds{R}+ & $\mathds{R}$ \\
    \verb+\mathbb{Q}+ & $\mathbb{Q}$ & 6 & \verb+\mathds{Q}+ & $\mathds{Q}$ \\
    \verb+\mathbb{N}+ & $\mathbb{N}$ & 6 & \verb+\mathds{N}+ & $\mathds{N}$ \\
    \bottomrule
    \end{tabular}
    \caption{32~classes which were not a single time classified correctly by
             the best CNN.}
    \label{table:hard-classes}
\end{table}

In contrast, 21~classes have an accuracy of more than \SI{99}{\percent} with
TF-CNN (see~\cref{table:easy-classes}).

\begin{table}[h]
    \centering
    \begin{tabular}{lcr}
    \toprule
    \LaTeX & Rendered & Total\\\midrule
    \verb+\forall        + & $\forall        $ & 214 \\
    \verb+\sim           + & $\sim           $ & 201 \\
    \verb+\nabla         + & $\nabla         $ & 122 \\
    \verb+\cup           + & $\cup           $ & 93  \\
    \verb+\neg           + & $\neg           $ & 85  \\
    \verb+\setminus      + & $\setminus      $ & 52  \\
    \verb+\supset        + & $\supset        $ & 42  \\
    \verb+\vdots         + & $\vdots         $ & 41  \\
    \verb+\boxtimes      + & $\boxtimes      $ & 22  \\
    \verb+\nearrow       + & $\nearrow       $ & 21  \\
    \verb+\uplus         + & $\uplus         $ & 19  \\
    \verb+\nvDash        + & $\nvDash        $ & 15  \\
    \verb+\AE            + & \AE               & 15  \\
    \verb+\Vdash         + & $\Vdash         $ & 14  \\
    \verb+\Leftarrow     + & $\Leftarrow     $ & 14  \\
    \verb+\upharpoonright+ & $\upharpoonright$ & 14  \\
    \verb+-              + & $-              $ & 12  \\
    \verb+\guillemotleft + & $\guillemotleft $ & 11  \\
    \verb+R              + & $R              $ & 9   \\
    \verb+7              + & $7              $ & 8   \\
    \verb+\blacktriangleright+ & $\blacktriangleright$ & 6 \\
    \bottomrule
    \end{tabular}
    \caption{21~classes with a class-wise accuracy of more than \SI{99}{\percent}
             with TF-CNN.}
    \label{table:easy-classes}
\end{table}

\section{Verification Challenge}
In the setting of an online symbol recognizer like
\href{http://write-math.com}{write-math.com} it is important to recognize when
the user enters a symbol which is not known to the classifier. One way to achieve
this is by training a binary classifier to recognize when two recordings belong to
the same symbol. This kind of task is similar to face verification.
Face verification is the task where two images with faces are given and it has
to be decided if they belong to the same person.

For the verification challenge, a training-test split is given. The training
data contains images with their class labels. The test set
contains 32~symbols which were not seen by the classifier before. The elements
of the test set are pairs of recorded handwritten symbols $(r_1, r_2)$. There
are three groups of tests:
\begin{enumerate}[label=V\arabic*]
    \item $r_1$ and $r_2$ both belong to symbols which are in the training set,
    \item $r_1$ belongs to a symbol in the training set, but $r_2$
          might not
    \item $r_1$ and $r_2$ don't belong symbols in the training set
\end{enumerate}

When evaluating models, the models may not take advantage of the fact that it
is known if a recording $r_1$ / $r_2$ is an instance of the training symbols.
For all test sets, the following numbers should be reported: True Positive (TP),
True Negative (TN), False Positive (FP), False Negative (FN),
Accuracy $= \frac{TP+ TN}{TP+TN+FP+FN}$.




\section{Acknowledgment}

I want to thank \enquote{Begabtenstiftung Informatik Karls\-ruhe}, the Foundation
for Gifted Informatics Students in Karlsruhe. Their support helped me to write
this work.

\bibliographystyle{IEEEtranSA}
\bibliography{literatur}

\appendix
\section*{Obtaining the data}
The data can be found at \dbDownloadURL. It is a \verb+tar.gz+ file of
\SI{\dbSizeMB}{\mega\byte}. The file can be verified with the MD5sum

\texttt{\dbMDfivesum}

The data is published under the ODbL~license. If you use
the \dbName~dataset, please cite this paper.

The \verb+tar.gz+ archive contains all data as png images and CSV files with
labels. The CSV files have the
columns \verb+path,symbol_id,latex,user_id+ with a header row. The \verb+path+ is the
relative path to a training example to the CSV file, e.g. \verb+../hasy-data/v2-00000.png+. The
\verb+symbol_id+ is an internal numeric identifier for the symbol class. The
website \href{http://write-math.com/symbol/?id=968}{write-math.com/symbol/?id=[symbol\_id]}
gives information related to the symbol. The column \verb+latex+ contains the
\LaTeX{} command associated with the class.
\onecolumn
\section*{Symbol Classes}
            \begin{longtable}{lc|lc}
                \toprule
                \LaTeX & Rendered & \LaTeX & Rendered \\
                \midrule
                \endhead
                \hline \multicolumn{4}{r}{{Continued on next page}} \\
                \endfoot
                \bottomrule
                \caption{112 symbols of \dbName.}
                \endlastfoot
\verb+\&+ & $\&$ &\verb+\nmid+ & $\nmid$\\
\verb+\Im+ & $\Im$ &\verb+\nvDash+ & $\nvDash$\\
\verb+\Re+ & $\Re$ &\verb+\int+ & $\int$\\
\verb+\S+ & $\S$ &\verb+\fint+ & $\fint$\\
\verb+\Vdash+ & $\Vdash$ &\verb+\odot+ & $\odot$\\
\verb+\aleph+ & $\aleph$ &\verb+\oiint+ & $\oiint$\\
\verb+\amalg+ & $\amalg$ &\verb+\oint+ & $\oint$\\
\verb+\angle+ & $\angle$ &\verb+\varoiint+ & $\varoiint$\\
\verb+\ast+ & $\ast$ &\verb+\ominus+ & $\ominus$\\
\verb+\asymp+ & $\asymp$ &\verb+\oplus+ & $\oplus$\\
\verb+\backslash+ & $\backslash$ &\verb+\otimes+ & $\otimes$\\
\verb+\between+ & $\between$ &\verb+\parallel+ & $\parallel$\\
\verb+\blacksquare+ & $\blacksquare$ &\verb+\parr+ & $\parr$\\
\verb+\blacktriangleright+ & $\blacktriangleright$ &\verb+\partial+ & $\partial$\\
\verb+\bot+ & $\bot$ &\verb+\perp+ & $\perp$\\
\verb+\bowtie+ & $\bowtie$ &\verb+\pitchfork+ & $\pitchfork$\\
\verb+\boxdot+ & $\boxdot$ &\verb+\pm+ & $\pm$\\
\verb+\boxplus+ & $\boxplus$ &\verb+\prime+ & $\prime$\\
\verb+\boxtimes+ & $\boxtimes$ &\verb+\prod+ & $\prod$\\
\verb+\bullet+ & $\bullet$ &\verb+\propto+ & $\propto$\\
\verb+\checkmark+ & $\checkmark$ &\verb+\rangle+ & $\rangle$\\
\verb+\circ+ & $\circ$ &\verb+\rceil+ & $\rceil$\\
\verb+\circledR+ & $\circledR$ &\verb+\rfloor+ & $\rfloor$\\
\verb+\circledast+ & $\circledast$ &\verb+\rrbracket+ & $\rrbracket$\\
\verb+\circledcirc+ & $\circledcirc$ &\verb+\rtimes+ & $\rtimes$\\
\verb+\clubsuit+ & $\clubsuit$ &\verb+\sharp+ & $\sharp$\\
\verb+\coprod+ & $\coprod$ &\verb+\sphericalangle+ & $\sphericalangle$\\
\verb+\copyright+ & $\copyright$ &\verb+\sqcap+ & $\sqcap$\\
\verb+\dag+ & $\dag$ &\verb+\sqcup+ & $\sqcup$\\
\verb+\dashv+ & $\dashv$ &\verb+\sqrt{}+ & $\sqrt{}$\\
\verb+\diamond+ & $\diamond$ &\verb+\square+ & $\square$\\
\verb+\diamondsuit+ & $\diamondsuit$ &\verb+\star+ & $\star$\\
\verb+\div+ & $\div$ &\verb+\sum+ & $\sum$\\
\verb+\ell+ & $\ell$ &\verb+\times+ & $\times$\\
\verb+\flat+ & $\flat$ &\verb+\top+ & $\top$\\
\verb+\frown+ & $\frown$ &\verb+\triangle+ & $\triangle$\\
\verb+\guillemotleft+ & $\guillemotleft$ &\verb+\triangledown+ & $\triangledown$\\
\verb+\hbar+ & $\hbar$ &\verb+\triangleleft+ & $\triangleleft$\\
\verb+\heartsuit+ & $\heartsuit$ &\verb+\trianglelefteq+ & $\trianglelefteq$\\
\verb+\infty+ & $\infty$ &\verb+\triangleq+ & $\triangleq$\\
\verb+\langle+ & $\langle$ &\verb+\triangleright+ & $\triangleright$\\
\verb+\lceil+ & $\lceil$ &\verb+\uplus+ & $\uplus$\\
\verb+\lfloor+ & $\lfloor$ &\verb+\vDash+ & $\vDash$\\
\verb+\lhd+ & $\lhd$ &\verb+\varnothing+ & $\varnothing$\\
\verb+\lightning+ & $\lightning$ &\verb+\varpropto+ & $\varpropto$\\
\verb+\llbracket+ & $\llbracket$ &\verb+\vartriangle+ & $\vartriangle$\\
\verb+\lozenge+ & $\lozenge$ &\verb+\vdash+ & $\vdash$\\
\verb+\ltimes+ & $\ltimes$ &\verb+\with+ & $\with$\\
\verb+\mathds{1}+ & $\mathds{1}$ &\verb+\wp+ & $\wp$\\
\verb+\mathsection+ & $\mathsection$ &\verb+\wr+ & $\wr$\\
\verb+\mid+ & $\mid$ &\verb+\{+ & $\{$\\
\verb+\models+ & $\models$ &\verb+\|+ & $\|$\\
\verb+\mp+ & $\mp$ &\verb+\}+ & $\}$\\
\verb+\multimap+ & $\multimap$ &\verb+\vee+ & $\vee$\\
\verb+\nabla+ & $\nabla$ &\verb+\wedge+ & $\wedge$\\
\verb+\neg+ & $\neg$ &\verb+\barwedge+ & $\barwedge$
    \label{table:symbols-of-db-0}
    \end{longtable}

\begin{table}[ht]
        \centering

            \begin{tabular}{lc|lc|lc|lc}
                \toprule
                \LaTeX & Rendered & \LaTeX & Rendered & \LaTeX & Rendered & \LaTeX & Rendered \\
                \midrule
\verb+\#+ & $\#$ &\verb+A+ & $A$ &\verb+S+ & $S$ &\verb+i+ & $i$\\
\verb+\$+ & $\$$ &\verb+B+ & $B$ &\verb+T+ & $T$ &\verb+j+ & $j$\\
\verb+\%+ & $\%$ &\verb+C+ & $C$ &\verb+U+ & $U$ &\verb+k+ & $k$\\
\verb+++ & $+$ &\verb+D+ & $D$ &\verb+V+ & $V$ &\verb+l+ & $l$\\
\verb+-+ & $-$ &\verb+E+ & $E$ &\verb+W+ & $W$ &\verb+m+ & $m$\\
\verb+/+ & $/$ &\verb+F+ & $F$ &\verb+X+ & $X$ &\verb+n+ & $n$\\
\verb+0+ & $0$ &\verb+G+ & $G$ &\verb+Y+ & $Y$ &\verb+o+ & $o$\\
\verb+1+ & $1$ &\verb+H+ & $H$ &\verb+Z+ & $Z$ &\verb+p+ & $p$\\
\verb+2+ & $2$ &\verb+I+ & $I$ &\verb+[+ & $[$ &\verb+q+ & $q$\\
\verb+3+ & $3$ &\verb+J+ & $J$ &\verb+]+ & $]$ &\verb+r+ & $r$\\
\verb+4+ & $4$ &\verb+K+ & $K$ &\verb+a+ & $a$ &\verb+s+ & $s$\\
\verb+5+ & $5$ &\verb+L+ & $L$ &\verb+b+ & $b$ &\verb+u+ & $u$\\
\verb+6+ & $6$ &\verb+M+ & $M$ &\verb+c+ & $c$ &\verb+v+ & $v$\\
\verb+7+ & $7$ &\verb+N+ & $N$ &\verb+d+ & $d$ &\verb+w+ & $w$\\
\verb+8+ & $8$ &\verb+O+ & $O$ &\verb+e+ & $e$ &\verb+x+ & $x$\\
\verb+9+ & $9$ &\verb+P+ & $P$ &\verb+f+ & $f$ &\verb+y+ & $y$\\
\verb+<+ & $<$ &\verb+Q+ & $Q$ &\verb+g+ & $g$ &\verb+z+ & $z$\\
\verb+>+ & $>$ &\verb+R+ & $R$ &\verb+h+ & $h$ &\verb+|+ & $|$\\

        \bottomrule
    \end{tabular}

    \caption{72 ASCII symbols of \dbName, including all
             ten digits, the Latin alphabet in lower and upper case and
             a few more symbols.}
    \label{table:symbols-of-db-1}
\end{table}

\begin{table}[ht]
        \centering
            \begin{tabular}{lc|lc|lc}
                \toprule
                \LaTeX & Rendered & \LaTeX & Rendered & \LaTeX & Rendered \\
                \midrule
\verb+\approx+ & $\approx$ &\verb+\geqslant+ & $\geqslant$ &\verb+\lesssim+ & $\lesssim$\\
\verb+\doteq+ & $\doteq$ &\verb+\neq+ & $\neq$ &\verb+\backsim+ & $\backsim$\\
\verb+\simeq+ & $\simeq$ &\verb+\not\equiv+ & $\not\equiv$ &\verb+\sim+ & $\sim$\\
\verb+\equiv+ & $\equiv$ &\verb+\preccurlyeq+ & $\preccurlyeq$ &\verb+\succ+ & $\succ$\\
\verb+\geq+ & $\geq$ &\verb+\preceq+ & $\preceq$ &\verb+\prec+ & $\prec$\\
\verb+\leq+ & $\leq$ &\verb+\succeq+ & $\succeq$ &\verb+\gtrless+ & $\gtrless$\\
\verb+\leqslant+ & $\leqslant$ &\verb+\gtrsim+ & $\gtrsim$ &\verb+\cong+ & $\cong$\\
        \bottomrule
    \end{tabular}

    \caption{21 symbols which are in \dbName and indicate a relationship.}
    \label{table:symbols-of-db-2}
\end{table}

\begin{table}[ht]
        \centering

            \begin{tabular}{lc|lc}
                \toprule
                \LaTeX & Rendered & \LaTeX & Rendered \\
                \midrule
\verb+\Downarrow+ & $\Downarrow$ &\verb+\nrightarrow+ & $\nrightarrow$\\
\verb+\Leftarrow+ & $\Leftarrow$ &\verb+\rightarrow+ & $\rightarrow$\\
\verb+\Leftrightarrow+ & $\Leftrightarrow$ &\verb+\rightleftarrows+ & $\rightleftarrows$\\
\verb+\Longleftrightarrow+ & $\Longleftrightarrow$ &\verb+\rightrightarrows+ & $\rightrightarrows$\\
\verb+\Longrightarrow+ & $\Longrightarrow$ &\verb+\rightsquigarrow+ & $\rightsquigarrow$\\
\verb+\Rightarrow+ & $\Rightarrow$ &\verb+\searrow+ & $\searrow$\\
\verb+\circlearrowleft+ & $\circlearrowleft$ &\verb+\shortrightarrow+ & $\shortrightarrow$\\
\verb+\circlearrowright+ & $\circlearrowright$ &\verb+\twoheadrightarrow+ & $\twoheadrightarrow$\\
\verb+\curvearrowright+ & $\curvearrowright$ &\verb+\uparrow+ & $\uparrow$\\
\verb+\downarrow+ & $\downarrow$ &\verb+\rightharpoonup+ & $\rightharpoonup$\\
\verb+\hookrightarrow+ & $\hookrightarrow$ &\verb+\rightleftharpoons+ & $\rightleftharpoons$\\
\verb+\leftarrow+ & $\leftarrow$ &\verb+\longmapsto+ & $\longmapsto$\\
\verb+\leftrightarrow+ & $\leftrightarrow$ &\verb+\mapsfrom+ & $\mapsfrom$\\
\verb+\longrightarrow+ & $\longrightarrow$ &\verb+\mapsto+ & $\mapsto$\\
\verb+\nRightarrow+ & $\nRightarrow$ &\verb+\leadsto+ & $\leadsto$\\
\verb+\nearrow+ & $\nearrow$ &\verb+\upharpoonright+ & $\upharpoonright$\\

        \bottomrule
    \end{tabular}

    \caption{32 arrow symbols of \dbName.}
    \label{table:symbols-of-db-3}
\end{table}

\begin{table}[ht]
        \centering

            \begin{tabular}{lc|lc|lc}
                \toprule
                \LaTeX & Rendered & \LaTeX & Rendered & \LaTeX & Rendered \\
                \midrule
\verb+\alpha+ & $\alpha$ &\verb+\xi+ & $\xi$ &\verb+\Xi+ & $\Xi$\\
\verb+\beta+ & $\beta$ &\verb+\pi+ & $\pi$ &\verb+\Pi+ & $\Pi$\\
\verb+\gamma+ & $\gamma$ &\verb+\rho+ & $\rho$ &\verb+\Sigma+ & $\Sigma$\\
\verb+\delta+ & $\delta$ &\verb+\sigma+ & $\sigma$ &\verb+\Phi+ & $\Phi$\\
\verb+\epsilon+ & $\epsilon$ &\verb+\tau+ & $\tau$ &\verb+\Psi+ & $\Psi$\\
\verb+\zeta+ & $\zeta$ &\verb+\phi+ & $\phi$ &\verb+\Omega+ & $\Omega$\\
\verb+\eta+ & $\eta$ &\verb+\chi+ & $\chi$ &\verb+\varepsilon+ & $\varepsilon$\\
\verb+\theta+ & $\theta$ &\verb+\psi+ & $\psi$ &\verb+\varkappa+ & $\varkappa$\\
\verb+\iota+ & $\iota$ &\verb+\omega+ & $\omega$ &\verb+\varpi+ & $\varpi$\\
\verb+\kappa+ & $\kappa$ &\verb+\Gamma+ & $\Gamma$ &\verb+\varrho+ & $\varrho$\\
\verb+\lambda+ & $\lambda$ &\verb+\Delta+ & $\Delta$ &\verb+\varphi+ & $\varphi$\\
\verb+\mu+ & $\mu$ &\verb+\Theta+ & $\Theta$ &\verb+\vartheta+ & $\vartheta$\\
\verb+\nu+ & $\nu$ &\verb+\Lambda+ & $\Lambda$ &\verb+ + & $ $\\

        \bottomrule
    \end{tabular}

    \caption{All Greek letters and some variations of Greek letters were
             used for evaluation. 38 of them are in this table, the rest
             is identical to Latin letters.}
    \label{table:symbols-of-db-4}
\end{table}

\begin{table}[ht]
        \centering

            \begin{tabular}{lc|lc|lc}
                \toprule
                \LaTeX & Rendered & \LaTeX & Rendered & \LaTeX & Rendered \\
                \midrule
\verb+\mathcal{A}+ & $\mathcal{A}$ &\verb+\mathcal{T}+ & $\mathcal{T}$ &\verb+\mathds{Z}+ & $\mathds{Z}$\\
\verb+\mathcal{B}+ & $\mathcal{B}$ &\verb+\mathcal{U}+ & $\mathcal{U}$ &\verb+\mathfrak{A}+ & $\mathfrak{A}$\\
\verb+\mathcal{C}+ & $\mathcal{C}$ &\verb+\mathcal{X}+ & $\mathcal{X}$ &\verb+\mathfrak{M}+ & $\mathfrak{M}$\\
\verb+\mathcal{D}+ & $\mathcal{D}$ &\verb+\mathcal{Z}+ & $\mathcal{Z}$ &\verb+\mathfrak{S}+ & $\mathfrak{S}$\\
\verb+\mathcal{E}+ & $\mathcal{E}$ &\verb+\mathbb{H}+ & $\mathbb{H}$ &\verb+\mathfrak{X}+ & $\mathfrak{X}$\\
\verb+\mathcal{F}+ & $\mathcal{F}$ &\verb+\mathbb{N}+ & $\mathbb{N}$ &\verb+\mathscr{A}+ & $\mathscr{A}$\\
\verb+\mathcal{G}+ & $\mathcal{G}$ &\verb+\mathbb{Q}+ & $\mathbb{Q}$ &\verb+\mathscr{C}+ & $\mathscr{C}$\\
\verb+\mathcal{H}+ & $\mathcal{H}$ &\verb+\mathbb{R}+ & $\mathbb{R}$ &\verb+\mathscr{D}+ & $\mathscr{D}$\\
\verb+\mathcal{L}+ & $\mathcal{L}$ &\verb+\mathbb{Z}+ & $\mathbb{Z}$ &\verb+\mathscr{E}+ & $\mathscr{E}$\\
\verb+\mathcal{M}+ & $\mathcal{M}$ &\verb+\mathds{C}+ & $\mathds{C}$ &\verb+\mathscr{F}+ & $\mathscr{F}$\\
\verb+\mathcal{N}+ & $\mathcal{N}$ &\verb+\mathds{E}+ & $\mathds{E}$ &\verb+\mathscr{H}+ & $\mathscr{H}$\\
\verb+\mathcal{O}+ & $\mathcal{O}$ &\verb+\mathds{N}+ & $\mathds{N}$ &\verb+\mathscr{L}+ & $\mathscr{L}$\\
\verb+\mathcal{P}+ & $\mathcal{P}$ &\verb+\mathds{P}+ & $\mathds{P}$ &\verb+\mathscr{P}+ & $\mathscr{P}$\\
\verb+\mathcal{R}+ & $\mathcal{R}$ &\verb+\mathds{Q}+ & $\mathds{Q}$ &\verb+\mathscr{S}+ & $\mathscr{S}$\\
\verb+\mathcal{S}+ & $\mathcal{S}$ &\verb+\mathds{R}+ & $\mathds{R}$ &\verb+ + & $ $\\

        \bottomrule
    \end{tabular}

    \caption{44 variants of Latin letters in \dbName.}
    \label{table:symbols-of-db-5}
\end{table}

\begin{table}[ht]
        \centering

            \begin{tabular}{lc|lc|lc}
                \toprule
                \LaTeX & Rendered & \LaTeX & Rendered & \LaTeX & Rendered \\
                \midrule
\verb+\therefore+ & $\therefore$ &\verb+\cdot+  & $\cdot$  &\verb+\dots+  & $\dots$\\
\verb+\because+   & $\because$   &\verb+\vdots+ & $\vdots$ &\verb+\ddots+ & $\ddots$\\
\verb+\dotsc+     & $\dotsc$     &\verb+ +      &          &\verb+ +      &  \\
        \bottomrule
    \end{tabular}

    \caption{7 symbols that contain only dots in \dbName.}
    \label{table:symbols-of-db-6}
\end{table}

\begin{table}[ht]
        \centering
            \begin{tabular}{lc|lc|lc|lc|lc}
                \toprule
                \LaTeX & R & \LaTeX & R & \LaTeX & R &  \LaTeX & R & \LaTeX & R \\
                \midrule
\verb+\AA+ & {\r A}                                                           &\verb+\L+       & \includegraphics[height=12.3pt, keepaspectratio]{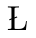} &\verb+\male+     & {\mbox {\wasyfamily \char 26}}                         &\verb+\ohm+      &  $\Omega $                     &\verb+\sun+       & {\mbox {\wasyfamily \char 46}} \\
\verb+\AE+ & \includegraphics[height=12.3pt, keepaspectratio]{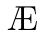} &\verb+\O+       & \includegraphics[height=12.3pt, keepaspectratio]{symbols/L.pdf} &\verb+\mars+     & {\leavevmode \lower 0.2ex\hbox {\wasyfamily \char 26}} &\verb+\fullmoon+ & {\mbox {\wasyfamily \char 35}} &\verb+\degree+    & {\ensuremath {^\circ }}\\
\verb+\aa+ & {\r a}                                                           &\verb+\o+       & \includegraphics[height=12.3pt, keepaspectratio]{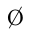} &\verb+\female+   & {\mbox {\wasyfamily \char 25}}                         &\verb+\leftmoon+ & {\mbox {\wasyfamily \char 36}} &\verb+\iddots+    & \includegraphics[height=12.3pt, keepaspectratio]{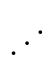}\\
\verb+\ae+ & \includegraphics[height=12.3pt, keepaspectratio]{symbols/AE.pdf} &\verb+\Bowtie+  & {\mbox {\wasyfamily \char 49}}                                  &\verb+\venus+    & {\leavevmode \raise 0.2ex\hbox {\wasyfamily \char 25}} &\verb+\checked+  & {\mbox {\wasyfamily \char 8}}  &\verb+\diameter+  & {\mbox {\wasyfamily \char 31}} \\
\verb+\ss+ & \includegraphics[height=12.3pt, keepaspectratio]{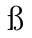} &\verb+\celsius+ & $^\circ \mathrm {C}$                                            &\verb+\astrosun+ & {\mbox {$\odot $}}                                     &\verb+\pounds+   & \textsterling                  &\verb+\mathbb{1}+ & \includegraphics[height=12.3pt, keepaspectratio]{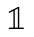}\\
        \bottomrule
    \end{tabular}
    \caption{25 symbols of \dbName.}
    \label{table:symbols-of-db-7}
\end{table}

\begin{table}[ht]
        \centering

            \begin{tabular}{lc|lc|lc}
                \toprule
                \LaTeX & Rendered & \LaTeX & Rendered & \LaTeX & Rendered \\
                \midrule
\verb+\cup+ & $\cup$ &\verb+\varsubsetneq+ & $\varsubsetneq$ &\verb+\exists+ & $\exists$\\
\verb+\cap+ & $\cap$ &\verb+\nsubseteq+ & $\nsubseteq$ &\verb+\nexists+ & $\nexists$\\
\verb+\emptyset+ & $\emptyset$ &\verb+\sqsubseteq+ & $\sqsubseteq$ &\verb+\forall+ & $\forall$\\
\verb+\setminus+ & $\setminus$ &\verb+\subseteq+ & $\subseteq$ &\verb+\in+ & $\in$\\
\verb+\supset+ & $\supset$ &\verb+\subsetneq+ & $\subsetneq$ &\verb+\ni+ & $\ni$\\
\verb+\subset+ & $\subset$ &\verb+\supseteq+ & $\supseteq$ &\verb+\notin+ & $\notin$\\

        \bottomrule
    \end{tabular}

    \caption{18 set related symbols of \dbName.}
    \label{table:symbols-of-db-8}
\end{table}
\clearpage
\twocolumn

\end{document}